# Does Machine Learning Outperform Traditional Fibrosis Scores in Predicting Liver Cirrhosis Risk? A Longitudinal EHR-Based Study

Zhuqi Miao[a,b], Ahmed G Qasem[c], Sujan Ravi[d], Jason T. Cheng[e], Courtney W. Houchen[e], Sumayah Abed[f], Dilorom Azimdjanovna Zuparova[c], Abdulaziz Ahmed[c]*

[a]Center for Health Systems Innovation, Oklahoma State University, Tulsa, OK, 74119, USA
[b]Department of Management Science and Information Systems, Oklahoma State University, Stillwater, OK, 74106, USA
[c]Department of Health Services Administration, University of Alabama at Birmingham, Birmingham, AL, 35233, USA
[d]Division of Gastroenterology and Hepatology, Department of Medicine, University of Alabama at Birmingham, Birmingham, AL, 35233, USA.
[e]College of Medicine, The University of Oklahoma Health Campus, Oklahoma City, OK, 73104, USA.
[f]Department of Family and Community Medicine, University of Alabama at Birmingham, Birmingham, AL, 35233, USA.
*Corresponding Author: Abdulaziz Ahmed, Email: aahmed2@uab.edu

## Abstract

**Objective:** Develop and evaluate machine learning (ML) models for predicting incident liver cirrhosis (LC) one and two years before diagnosis using routinely collected electronic health record (EHR) data and compare their performance with the FIB-4 and APRI clinical scores.

**Methods:** We conducted a retrospective cohort study using de-identified EHR data from a large academic health system. Adult patients with diagnostic evidence of LC or LC-related risk conditions were identified using ICD-9/10 codes and classified into cirrhosis and non-cirrhosis cohorts. One- and two-year prediction scenarios were created using observation and prediction windows. Demographics, diagnoses, laboratory results, and vital signs from the observation window were used as predictors. XGBoost models were developed with feature selection and Bayesian hyperparameter tuning and evaluated on held-out test sets. The performance of XGBoost, FIB-4, and APRI were compared on the same test data using accuracy, precision, recall, F1 score, AUC, and PR AUC.

**Results:** The final cohorts included 54,365 patients for the 1-year prediction and 43,743 for the 2-year prediction. XGBoost consistently outperformed FIB-4 and APRI across both prediction horizons. The ML models achieved AUCs of 0.834 and 0.811 versus 0.700 and 0.677 for FIB-4 and 0.744 and 0.719 for APRI. PR AUCs were 0.502 and 0.434 for XGBoost compared with 0.310 and 0.241 for FIB-4 and 0.372 and 0.306 for APRI.

**Conclusions:** ML models using routine EHR data substantially outperform traditional clinical scores for early LC prediction, enabling more accurate risk stratification and supporting earlier clinical intervention through automated decision support.

# Introduction

Liver cirrhosis (LC) is the final stage of chronic liver disease and a leading cause of morbidity and mortality worldwide.[1] It develops silently over years, often without symptoms until the disease has progressed to an advanced stage.[2,3] Patients with liver cirrhosis are associated with reduced survival, increased hospitalization rates, and significant economic burden. In the United States, chronic liver disease and cirrhosis contribute a substantial share of the national health burden. Mortality from cirrhosis rose by nearly 90 percent over the past three decades, increasing from about 35,000 deaths in 1990 to more than 65,000 in 2019. Cirrhosis ranked as the 11th leading cause of death in 2019 and accounted for over 2 percent of all deaths.[3]

Early identification of patients at high risk is critical, because timely intervention can slow progression, guide treatment decisions, and prevent complications such as hepatic decompensation, hepatocellular carcinoma, and death.[4,5] Despite the importance of accurate staging, current tools for routine risk assessment have limitations. Liver biopsy remains the gold standard for diagnosing liver disease, characterizing masses, and grading inflammation, fibrosis, and steatosis. It provides detailed histology but carries procedural risks, requires specialized expertise, and is not suitable for repeated monitoring.[6,7] Newer approaches such as liquid biopsy and digital pathology are emerging and may reduce reliance on conventional biopsy, although their sensitivity and specificity remain suboptimal.[8] Over the past decades, non-invasive tests have become more widely used to assess liver disease. FibroScan (transient elastography) is considered the most reliable non-invasive method for evaluating liver fibrosis and is widely accepted, provides rapid results, and demonstrates high patient acceptability. However, it is limited by high costs and lack of availability in smaller cities, confining its routine use to tertiary care centers and specialty hospitals.[9]

To address these limitations, several serum-based indices have been developed, including the Fibrosis-4 (FIB-4),[10] Fibrofast (FIB-5) scores,[11,12] and AST-to-Platelet Ratio Index (APRI). These formulas rely on laboratory and demographic variables that are routinely available in clinical practice. FIB-4 is derived from age, platelet count, aspartate aminotransferase (AST), and alanine aminotransferase (ALT). FIB-5 incorporates albumin, alkaline phosphatase, platelet count, and the AST to ALT ratio. On the other hand, APRI is calculated from AST scaled by its upper limit of normal and the platelet count. These indices are inexpensive to calculate, require no specialized equipment, and can be applied in primary care settings. Evidence supports the use of these indices as rule-out tools for advanced fibrosis. In a cohort of patients with non-alcoholic fatty liver disease, Kumari et al.[11] reported that FIB-4 achieved an area under the receiver operating characteristic (ROC) curve (AUC) of 0.712 and FIB-5 achieved an AUC of 0.655 for detecting advanced fibrosis. Both indices showed negative predictive values above 90 percent, meaning that patients classified as low risk were almost always free of advanced disease. However, their ability to detect advanced fibrosis remained limited. At the <2.02 cutoff, FIB-4 identified only 46.8 percent of patients with advanced fibrosis (sensitivity, true positives) while correctly classifying 86.4 percent of patients without advanced fibrosis (specificity, true negatives). This means that more than half of patients with advanced disease were missed (false negatives). At the <-7.11 cutoff, FIB-5 identified 81.8 percent of patients with advanced fibrosis (sensitivity) but correctly excluded only 46.8 percent of those without advanced disease (specificity), leading to more than half of patients without advanced fibrosis being incorrectly classified as having advanced fibrosis (false positives). These tradeoffs indicate that while both scores are useful for excluding advanced fibrosis, they are not sufficiently reliable for accurate risk stratification across diverse populations.[13,14]

The limited diagnostic performance of traditional serum-based scores reflects their structural simplicity. Both indices apply fixed weights and simple arithmetic combination to a small set of laboratory parameters, without accounting for sophisticated nonlinear associations or the wider set of demographics, biochemical, and clinical variables that may contribute to disease progression. Fibrosis and cirrhosis arise from complex, multifactorial processes involving metabolic dysfunction, chronic inflammation, genetic predisposition, and comorbid conditions. Static formulas of this kind are unable to adapt to these patterns. Consequently, while they are useful for screening and for ruling out advanced fibrosis in low-risk populations, they lack the precision required for clinical decision support in contexts where accurate staging is essential for treatment and referral.

Machine learning (ML) offers a potential solution to these limitations. Unlike traditional scores, ML algorithms are data-driven and capable of integrating a broader range of predictors. They can capture complex feature interactions and improve performance as larger datasets become available. In hepatology and related fields, ML approaches have been applied successfully to predict outcomes including hepatocellular carcinoma, portal hypertension, and mortality, with performance generally exceeding that of conventional regression-based risk scores. These findings suggest that ML is well suited to enhance the prediction of cirrhosis, particularly when benchmarked directly against indices such as FIB-4 and APRI. Accordingly, this study develops and evaluates longitudinal EHR-based XGBoost models to predict incident LC one and two years before diagnosis and directly compares their performance with FIB-4 and APRI.

# Methods

## Data Source and Extraction

This study utilized de-identified electronic health record (EHR) data from the University of Alabama at Birmingham (UAB) Informatics for Integrating Biology and the Bedside (i2b2) database to identify candidate patients. The UAB-i2b2 database contains structured inpatient and outpatient clinical records derived from the enterprise EHR system and clinical data warehouse, encompassing encounters across the UAB healthcare system. Data were obtained for encounters occurring between January 1, 2014, and December 31, 2024. All data were de-identified in compliance with the Health Insurance Portability and Accountability Act (HIPAA), and the study was exempted from Institutional Review Board (IRB) review at UAB.

For cohort extraction, adult candidate patients were identified based on diagnostic evidence of LC or established LC-related risk conditions using ICD-9-CM and ICD-10-CM diagnosis codes. Relevant LC risk conditions included metabolic dysfunction–associated steatotic liver disease (MASLD), metabolic dysfunction–associated steatohepatitis (MASH), alcohol-related liver disease, viral hepatitis, autoimmune and cholestatic liver diseases, hereditary liver diseases, toxic or drug-induced liver disease, congestive hepatopathy, and metabolic comorbidities (Table 1).[15] Patients younger than 18 years were excluded.

The candidate pool was then classified into two cohorts. Case patients were defined as individuals with an LC diagnosis based on ICD-9-CM codes 571.2 or 571.5 or ICD-10-CM codes K74.6x or K70.3x. Control patients were defined as individuals with one or more LC-related risk conditions but no recorded LC diagnosis code. For both cohorts, candidate predictors were extracted across four domains: Eight demographic characteristics, clinically grouped diagnosis indicators corresponding to LC-related risk conditions listed in Table 1, nine laboratory tests identified using Logical Observation Identifiers Names and Codes (LOINC), and three vital signs. A detailed list of extracted variables is provided in Table 1.

Table 1: Diagnosis Code Definitions for Cohort Extraction and Candidate Predictors Used for Model Development

| **Clinical condition** | **Description** |
|---|---|
| LC-related risk conditions | ICD-9/10 codes for MASLD/MASH, obesity, type 2 diabetes, dyslipidemia, metabolic syndrome, alcohol use disorder, alcohol-related liver disease, chronic viral hepatitis B and C, autoimmune hepatitis, primary biliary cholangitis, primary sclerosing cholangitis, hereditary hemochromatosis, Wilson disease, alpha-1 antitrypsin deficiency, drug-induced or toxic liver disease, and congestive hepatopathy. |
| Liver Cirrhosis | |
| 571.2 | Alcoholic cirrhosis of liver |
| 571.5 | Cirrhosis of liver without mention of alcohol |
| K70.3x | Alcoholic cirrhosis of liver |
| K74.6x | Other and unspecified cirrhosis of liver |
| **Predictor domain** | **Predictors** |
| Demographics | Date of birth, age, gender, race, marital status, county, state, and rural-urban status |
| Diagnoses | Clinically grouped indicators representing the LC-related risk conditions listed above |
| Laboratory Test Results (LOINC) | |
| 1742-6 | Alanine aminotransferase (ALT) |
| 1751-7 | Albumin |
| 1920-8 | Aspartate aminotransferase (AST) |
| 6768-6 | Alkaline phosphatase (ALP) |
| 1975-2 | Bilirubin, total in serum or plasma |
| 718-7 | Hemoglobin |
| 26515-7 | Platelets |
| 2885-2 | Protein, total |
| 5902-2 | Prothrombin time (PT) |
| Vital Signs | Systolic blood pressure (SBP)<br>Diastolic blood pressure (DBP)<br>Body mass index (BMI) |

## Data Preparation

Although the study was retrospective, the extracted data were organized to emulate a real-world prediction scenario.[16,17] Specifically, for each LC patient, the first recorded LC diagnosis date was identified and used as the index date. Two non-overlapping temporal windows were then defined relative to the index date, as illustrated in Figure 1:

- Prediction Window: spans one or two years immediately before the first LC diagnosis date. Data within this window were excluded from modeling to reduce the risk of information leakage from diagnostic evaluations, clinical workups, or disease-related encounters occurring close to diagnosis.
- Observation Window: immediately precedes the prediction window and extends as far back as available in the patient's EHR records. Data recorded during this window were aggregated and used for prediction model development.

This approach allows us to leverage all historical patient data before the prediction window to predict the onset of LC within one or two years. The two prediction window lengths were selected to align with the typical clinical timeline of periodic clinical checkups for patients at risk of LC.

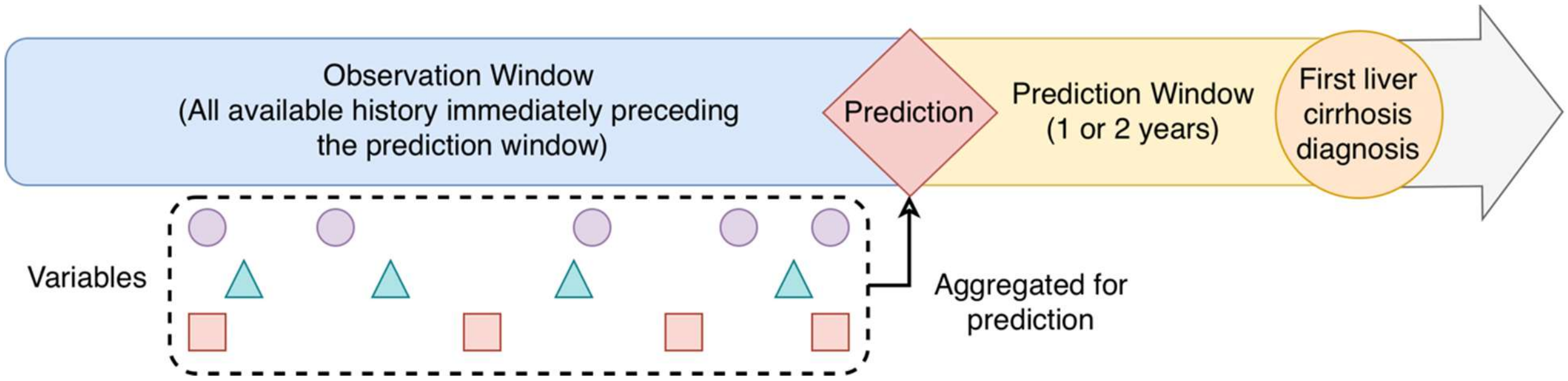

Figure 1. Illustration of data organization, with prediction and observation windows and data aggregation, used to emulate the prediction scenario.

To construct the non-LC cohort, we first selected patients from the candidate pool who never had LC. Additionally, in order to emulate a realistic recruitment scenario in which LC and non-LC patients are enrolled over similar time periods, we identified encounters from non-LC patients whose prediction times aligned with those of LC patients. Specifically, for each LC patient, we selected five non-LC patient encounters occurring within one week before or after that LC patient's prediction time. All selected non-LC encounters were then combined and filtered to retain only unique patients, as illustrated in Figure 2. This matching procedure was conducted separately for the 1-year and 2-year prediction horizons.

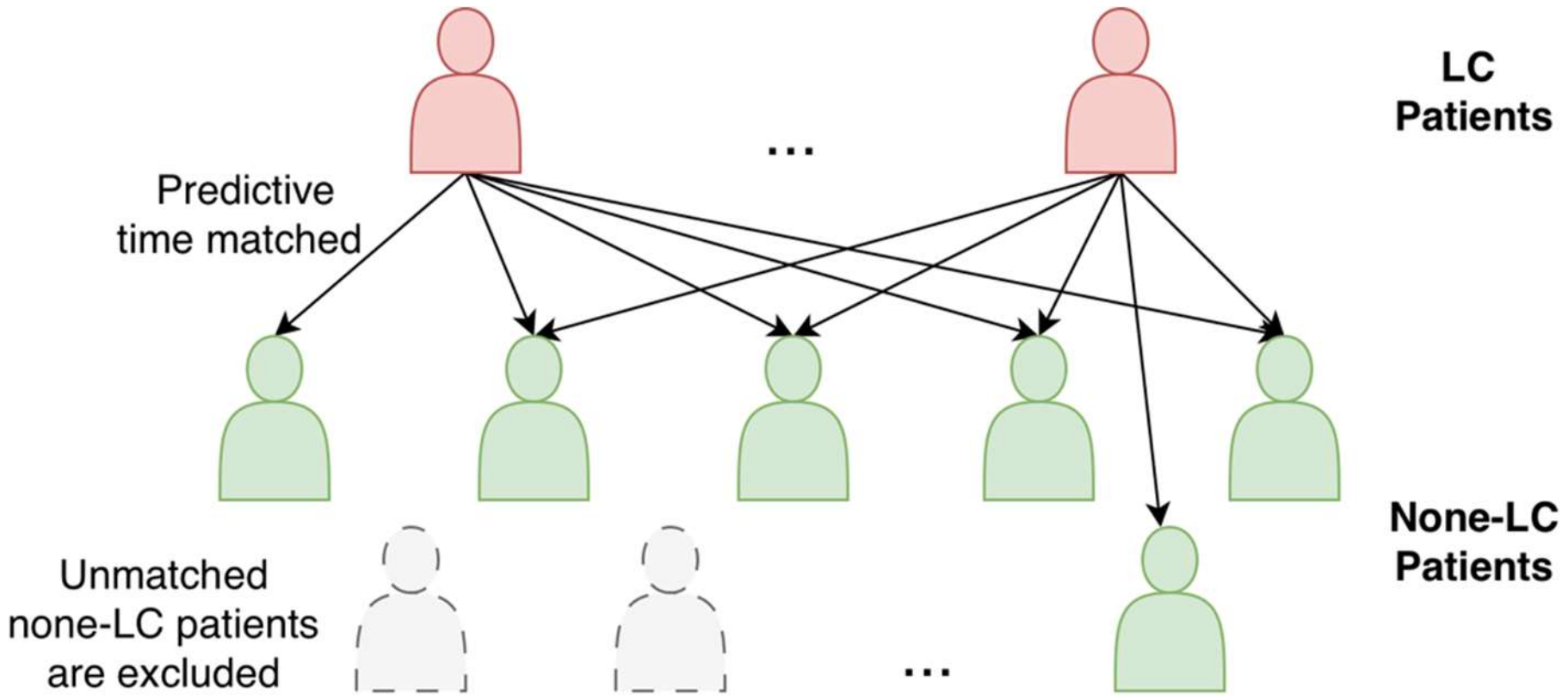

Figure 2. Illustration of the matching process. "Matched time" indicates that the encounters occurred no more than a week apart.

For data aggregation, patient-level predictors were constructed from the observation window. Diagnosis codes were mapped into clinically meaningful indicators representing LC-related risk conditions, rather than retaining the full high-dimensional ICD-9-CM and ICD-10-CM code set. The mapped diagnosis features captured MASLD, MASH, alcohol-related liver disease, chronic viral hepatitis, autoimmune and cholestatic liver diseases, hereditary liver diseases, drug-induced or toxic liver disease, congestive hepatopathy, toxic exposure, and metabolic comorbidities. Each diagnosis group was transformed into a binary variable based on the presence of any corresponding diagnosis code during the observation window. Specifically, a value of 1 was assigned if at least one mapped diagnosis code for that group occurred during the observation window; otherwise, a value of 0 was assigned if none of the mapped codes for that group were observed. Patients without a target code for a given diagnosis group were retained and assigned 0 for that indicator.

Numerical variables, including laboratory test results and vital signs, were averaged across the observation window. Demographic variables were defined at the prediction point, corresponding to one or two years before the first LC diagnosis date for cases, and the matched reference date for controls. Residential county and state information were linked to the 2023 Rural-Urban Continuum Codes (RUCC)[18] to derive rural-urban status. Categorical demographic variables were recoded to reduce sparse categories before one-hot encoding. Race was grouped as White, Black, or Other; marital status was grouped as Married, Single, Divorced, Widowed, or Other; and rural-urban status was retained as an ordinal variable.

Because EHR data are collected during routine care rather than according to a standardized research protocol, laboratory and vital-sign measurements were incomplete for some patients across the observation window. Missing values were not imputed before ML modeling, as the ML model we applied, XGBoost,[23] can accommodate missing predictor values during tree construction and learn how missingness contributes to prediction through its split structure.[19] This approach allowed the model to use all eligible patients with available EHR data rather than restricting model development to patients with complete laboratory panels. In contrast, FIB-4 and APRI were calculated only for patients with the laboratory values required for each score, since these serum-based indices cannot be computed when their component laboratory tests are missing. This distinction reflects an important practical advantage of flexible EHR-based ML models over traditional clinical scores in real-world data, where laboratory testing is often incomplete.

## Modeling

We developed separate XGBoost models for the 1-year and 2-year prediction horizons using the corresponding cleaned datasets. For each prediction horizon, the data were divided into training and testing sets using a 70/30 split stratified by LC status to preserve the case-control distribution in both partitions. The training set was used for feature selection, hyperparameter tuning, and final model development, while the held-out test set was reserved for final evaluation.

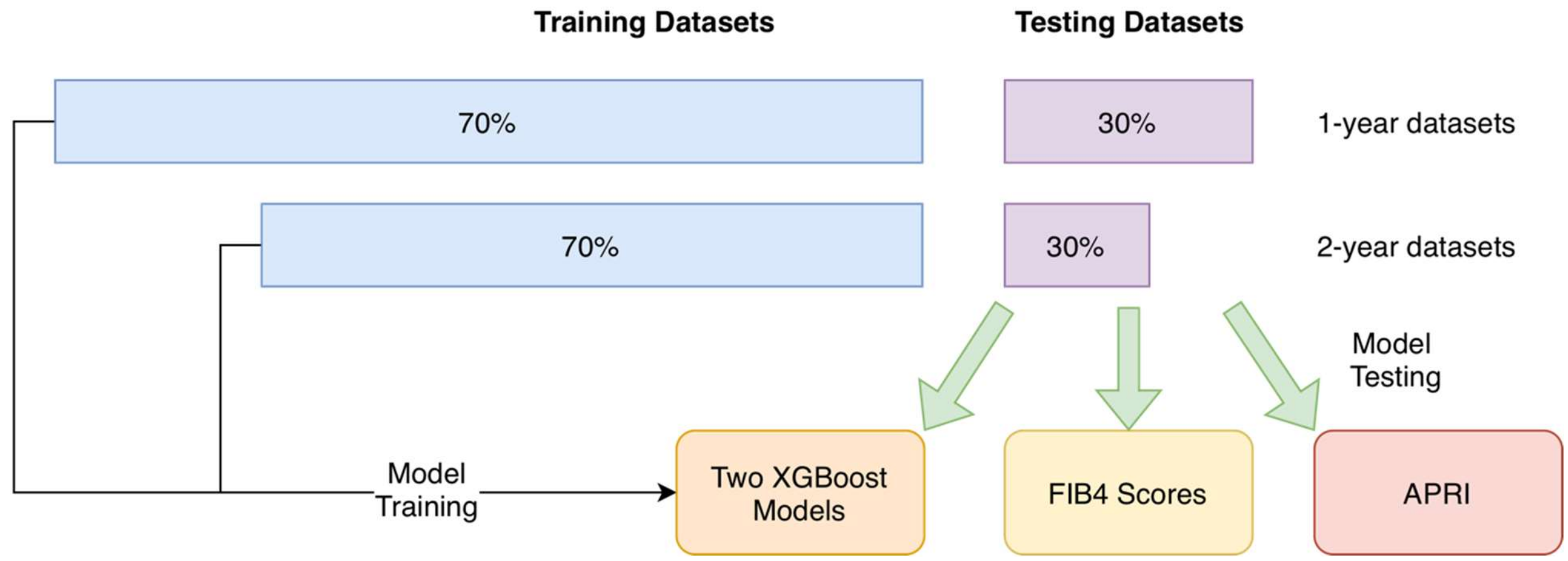

Figure 3. Schematic of the model training and testing process using multi-year datasets.

To reduce model complexity and improve generalizability, feature selection was performed within the training data before model tuning. An initial XGBoost classifier was fit using all eligible predictors, and feature importance scores from this model were used to rank candidate variables. The top-ranked predictors were then retained separately for each prediction horizon, resulting in 36 predictors for the 1-year model and 35 predictors for the 2-year model. This approach was used to reduce the influence of weak or redundant predictors, improve computational efficiency, and support a more parsimonious and clinically interpretable model. Feature selection was conducted using the training partition only to avoid information leakage from the test set.

After feature selection, Bayesian hyperparameter tuning was performed using Optuna to optimize model performance. Optuna is a hyperparameter optimization framework introduced by Akiba et al. Candidate hyperparameter combinations were evaluated within the training data using stratified 5-fold cross-validation, with mean validation ROC AUC used as the optimization criterion. The tuning process considered parameters related to model complexity, learning rate, sampling, and regularization, including the number of trees, maximum tree depth, learning rate, row and column subsampling, minimum child weight, gamma, and L1/L2 regularization terms. Within the training data, candidate hyperparameter combinations were evaluated using stratified 5-fold cross-validation, with mean validation AUC used as the optimization criterion. The final hyperparameter set for each prediction horizon was selected as the combination that achieved the highest mean cross-validation AUC. The search ranges and selected best values for the 1-year and 2-year models are summarized in Table 2, respectively. Using these selected parameters, each final XGBoost model was refitted on the full training set and subsequently evaluated on the held-out test set.

For comparison with established serum-based scores, FIB-4 and APRI were calculated for patients in the held-out test set who had complete laboratory values required for each score. XGBoost, FIB-4, and APRI were then evaluated on this same test subset to ensure a direct and consistent comparison across the three approaches. Performance was assessed using AUC, PR AUC, recall, precision, and F1-score.

Table 2. Hyperparameter tuning summary for the XGBoost prediction models

| **Hyperparameter** | **Search range considered** | **Best value selected** |
|---|---|---|
| XGBoost model for 1-year prediction | | |
| Number of trees (n_estimators) | 150–650 | 206 |
| Maximum tree depth (max_depth) | 3–8 | 6 |
| Learning rate (learning_rate) | 0.015–0.10 | 0.049 |
| Row subsampling (subsample) | 0.65–1.00 | 0.645 |
| Column subsampling (colsample_bytree) | 0.75–1.00 | 0.648 |
| Minimum child weight (min_child_weight) | 1–8 | 2 |
| Minimum loss reduction (gamma) | 0.00–3.00 | 1.856 |
| L1 regularization (reg_alpha) | 0.00–3.00 | 0.052 |
| L2 regularization (reg_lambda) | 0.50–8.00 | 0.662 |
| XGBoost model for 2-year prediction | | |
| Number of trees (n_estimators) | 120–320 | 253 |
| Maximum tree depth (max_depth) | 3–8 | 7 |
| Learning rate (learning_rate) | 0.01–0.15 | 0.103 |
| Row subsampling (subsample) | 0.60–1.00 | 0.932 |
| Column subsampling (colsample_bytree) | 0.60–1.00 | 0.873 |
| Minimum child weight (min_child_weight) | 1–10 | 3 |
| Minimum loss reduction (gamma) | 0.00–5.00 | 0.934 |
| L1 regularization (reg_alpha) | 0.00–5.00 | 2.243 |
| L2 regularization (reg_lambda) | 0.50–10.00 | 2.128 |

# Results

Figure 4 illustrates the cohort identification and filtering process. A total of 494,417 unique patients were identified from the candidate cohort, including 22,671 patients with LC and 471,746 patients without LC. After prediction-time matching, 78,693 unique non-LC controls were retained. The final cohort included 54,365 patients for the 1-year prediction window, consisting of 6,925 LC cases and 47,440 controls. For the 2-year prediction window, the final cohort included 43,743 patients, consisting of 5,418 LC cases and 38,325 controls.

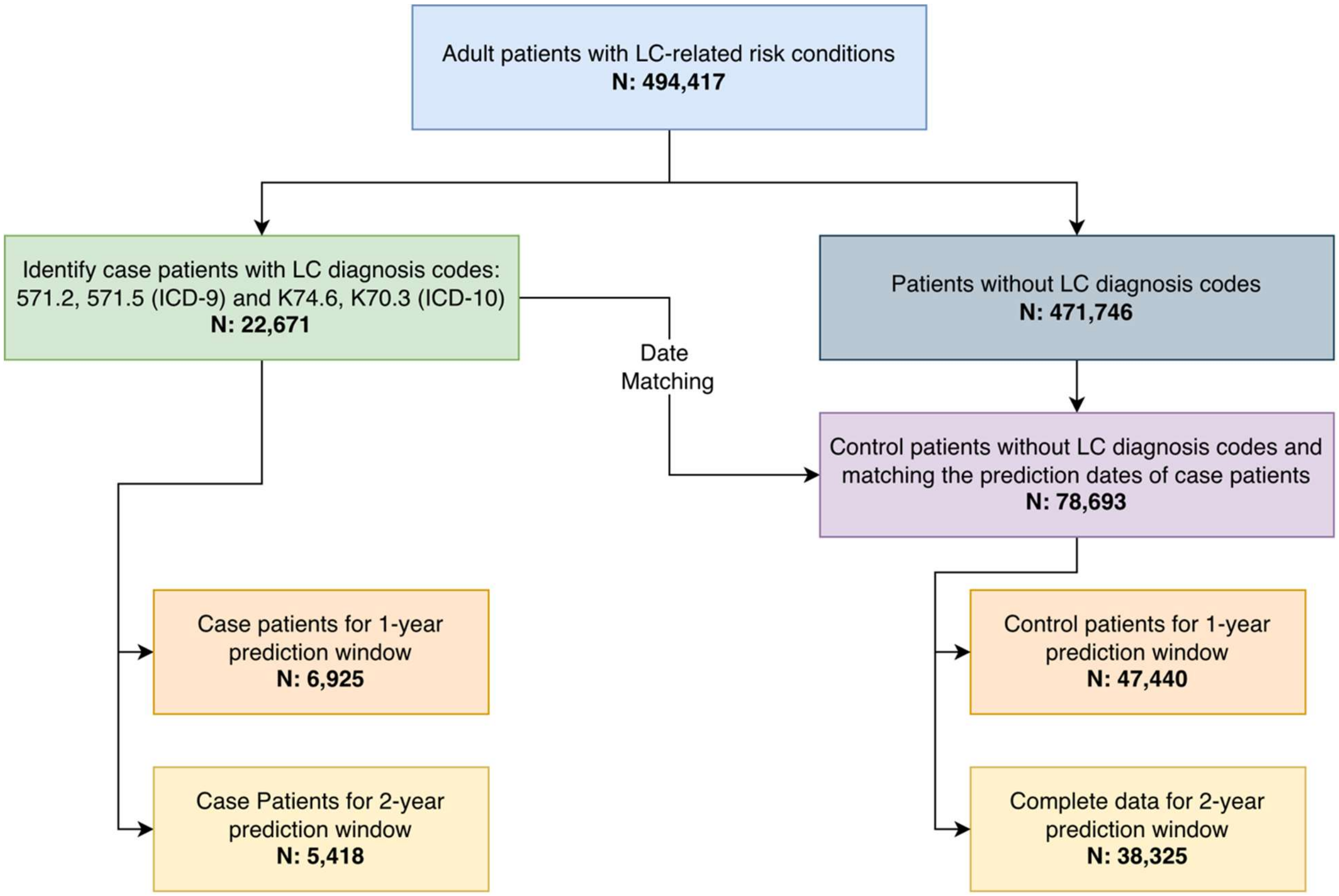

Figure 4. Flowchart showing the data extraction and cohort derivation process.

The cohort characteristics for the 1-year prediction window are presented in Table 3, while the corresponding 2-year characteristics are provided in Table A1 in the Appendix. Across both prediction horizons, several demographic characteristics differed significantly between LC cases and controls. In the 1-year cohort, LC cases were older than controls on average (57.7 vs 54.9 years) and were more frequently male (54.0% vs 39.3%), with similar patterns observed in the 2-year cohort (57.0 vs 54.3 years and 53.8% vs 39.1% male, respectively). Rural-urban status was also significantly associated with LC status in both cohorts ($p < 0.001$), with LC cases less frequently residing in large metropolitan counties and more frequently represented in smaller metropolitan and nonmetropolitan categories. For example, 62.1% of 1-year LC cases resided in counties within metropolitan areas of at least one million people, compared with 75.7% of controls; corresponding proportions in the 2-year cohort were 63.9% and 77.1%. These patterns demonstrate that demographic and geographic characteristics captured meaningful differences between LC and non-LC patients and may contribute to early risk stratification.

Laboratory and clinical profiles were also distinct between LC cases and controls. Across both prediction windows, LC cases had higher mean AST, ALT, total bilirubin, alkaline phosphatase, and total protein values, along with lower platelet counts and albumin levels, and these differences were statistically significant ($p < 0.001$). LC cases also had lower mean BMI and higher mean systolic and diastolic blood pressure than controls in both cohorts ($p < 0.001$). Hemoglobin did not differ significantly between cases and controls in either prediction window ($p = 0.513$ for the 1-year cohort and $p = 0.522$ for the 2-year cohort). Among LC-related risk indicators, liver-specific and etiologic conditions were generally more prevalent among LC cases, including chronic hepatitis C, chronic hepatitis B, MASLD/MASH, autoimmune hepatitis, primary biliary cholangitis, primary sclerosing cholangitis, alcohol use disorder, and alcohol-related liver disease (all $p < 0.001$ in both cohorts). In contrast, obesity, dyslipidemia/hyperlipidemia, and metabolic syndrome were less prevalent among LC cases than controls in both cohorts. Type 2 diabetes was similarly distributed between cases and controls and was not statistically significant in either cohort ($p = 0.729$ for the 1-year cohort and $p = 0.087$ for the 2-year cohort). Overall, the cohort characteristics demonstrate clinically meaningful differences between LC and non-LC patients and support the relevance of the demographic, laboratory, and diagnosis-based predictors included in the modeling datasets.

Table 3. Cohort characteristics for the 1-year prediction

| | **Overall** | **Controls** | **Cases** | **p-value** |
|---|---|---|---|---|
| **Patient Counts** | 54,365 | 47,440 | 6,925 | |
| **Gender** | | | | <0.001 |
| Female | 31,966 (58.8%) | 28,782 (60.7%) | 3,184 (46.0%) | |
| Male | 22,399 (41.2%) | 18,658 (39.3%) | 3,741 (54.0%) | |
| **Race** | | | | <0.001 |
| White | 31,099 (57.2%) | 26,478 (55.8%) | 4,621 (66.7%) | |
| Black | 20,321 (37.4%) | 18,344 (38.7%) | 1,977 (28.5%) | |
| Other | 2,945 (5.4%) | 2,618 (5.5%) | 327 (4.7%) | |
| **Marital Status** | | | | <0.001 |
| Married | 25,783 (47.4%) | 22,722 (47.9%) | 3,061 (44.2%) | |
| Single | 15,407 (28.3%) | 13,392 (28.2%) | 2,015 (29.1%) | |
| Divorced | 5,708 (10.5%) | 4,842 (10.2%) | 866 (12.5%) | |
| Widowed | 5,340 (9.8%) | 4,699 (9.9%) | 641 (9.3%) | |
| Other | 2,127 (3.9%) | 1,785 (3.8%) | 342 (4.9%) | |
| **Rural-Urban Status** | | | | <0.001 |
| Metro - Counties in metro areas of 1 million population or more | 40,209 (74.0%) | 35,907 (75.7%) | 4,302 (62.1%) | |
| Metro - Counties in metro areas of 250,000 to 1 million population | 4,401 (8.1%) | 3,581 (7.5%) | 820 (11.8%) | |
| Metro - Counties in metro areas of fewer than 250,000 population | 3,788 (7.0%) | 3,120 (6.6%) | 668 (9.6%) | |
| Nonmetro - Urban population of 20,000 or more, adjacent to a metro area | 2,632 (4.8%) | 2,147 (4.5%) | 485 (7.0%) | |
| Nonmetro - Urban population of 20,000 or more, not adjacent to a metro area | 256 (0.5%) | 209 (0.4%) | 47 (0.7%) | |
| Nonmetro - Urban population of 5,000 to 20,000, adjacent to a metro area | 1,293 (2.4%) | 1,043 (2.2%) | 250 (3.6%) | |
| Nonmetro - Urban population of 5,000 to 20,000, not adjacent to a metro area | 150 (0.3%) | 109 (0.2%) | 41 (0.6%) | |
| Nonmetro - Urban population of fewer than 5,000, adjacent to a metro area | 1,037 (1.9%) | 833 (1.8%) | 204 (2.9%) | |
| Nonmetro - Urban population of fewer than 5,000, not adjacent to a metro area | 599 (1.1%) | 491 (1.0%) | 108 (1.6%) | |
| **Age, mean (SD)** | 55.3 (15.9) | 54.9 (16.3) | 57.7 (13.3) | <0.001 |
| **Vital and Labs, mean (SD)** | | | | |
| BMI | 31.6 (7.8) | 31.7 (7.8) | 30.5 (7.7) | <0.001 |
| DBP | 79.1 (8.3) | 79.0 (8.2) | 79.6 (9.2) | <0.001 |
| SBP | 132.3 (15.2) | 132.0 (14.9) | 134.4 (16.7) | <0.001 |
| ALT | 25.8 (31.9) | 23.7 (26.8) | 41.3 (53.8) | <0.001 |
| Albumin | 4.0 (0.4) | 4.0 (0.4) | 3.9 (0.5) | <0.001 |
| AST | 26.5 (29.1) | 23.9 (24.6) | 45.6 (46.8) | <0.001 |
| Bilirubin | 0.6 (0.5) | 0.6 (0.4) | 0.9 (1.0) | <0.001 |
| Platelets | 237.2 (77.5) | 241.6 (75.2) | 205.0 (86.1) | <0.001 |
| Total Protein | 7.0 (0.6) | 7.0 (0.6) | 7.1 (0.7) | <0.001 |
| ALP | 81.6 (50.3) | 77.8 (37.1) | 109.0 (99.3) | <0.001 |
| Hemoglobin | 13.0 (1.8) | 13.0 (1.7) | 13.0 (2.0) | 0.513 |
| **LC-related risk indicators, Present, n(%)** | | | | |
| Alcohol-related liver disease | 167 (0.3%) | 50 (0.1%) | 117 (1.7%) | <0.001 |
| Alcohol use disorder | 2,586 (4.8%) | 1,872 (3.9%) | 714 (10.3%) | <0.001 |
| Alpha1 antitrypsin deficiency | 76 (0.1%) | 51 (0.1%) | 25 (0.4%) | <0.001 |
| Autoimmune hepatitis | 145 (0.3%) | 45 (0.1%) | 100 (1.4%) | <0.001 |
| Chronic hepatitis b | 258 (0.5%) | 119 (0.3%) | 139 (2.0%) | <0.001 |
| Chronic hepatitis c | 1,243 (2.3%) | 623 (1.3%) | 620 (9.0%) | <0.001 |
| Congestive hepatopathy | 148 (0.3%) | 91 (0.2%) | 57 (0.8%) | <0.001 |
| Drug-induced liver injury | 113 (0.2%) | 64 (0.1%) | 49 (0.7%) | <0.001 |
| Dyslipidemia hyperlipidemia | 20,720 (38.1%) | 18,959 (40.0%) | 1761 (25.4%) | <0.001 |
| Hereditary hemochromatosis | 40 (0.1%) | 28 (0.1%) | 12 (0.2%) | 0.002 |
| Masld-steatosis-mash | 3,285 (6.0%) | 2,366 (5.0%) | 919 (13.3%) | <0.001 |
| Metabolic syndrome | 935 (1.7%) | 849 (1.8%) | 86 (1.2%) | 0.001 |
| Obesity | 19,175 (35.3%) | 17,525 (36.9%) | 1,650 (23.8%) | <0.001 |
| Primary biliary cholangitis | 122 (0.2%) | 19 (0.0%) | 103 (1.5%) | <0.001 |
| Primary sclerosing cholangitis | 108 (0.2%) | 24 (0.1%) | 84 (1.2%) | <0.001 |
| Toxic exposure | 187 (0.3%) | 175 (0.4%) | 12 (0.2%) | 0.013 |
| Type2 diabetes | 14,019 (25.8%) | 12,221 (25.8%) | 1,798 (26.0%) | 0.729 |
| Wilson disease | 8 (0.0%) | 4 (0.0%) | 4 (0.1%) | 0.009 |

Model performance is summarized in Table 4 and illustrated in Figure 5. Across both prediction horizons, the tuned XGBoost models consistently outperformed FIB-4 and APRI on AUC, PR AUC, recall, precision, and F1-score when the three approaches were evaluated on the same test subset. For the 1-year prediction window, XGBoost achieved an AUC of 0.834 and a PR AUC of 0.502, compared with 0.700 and 0.310 for FIB-4 and 0.744 and 0.372 for APRI, respectively. This corresponded to improvements of 0.134 in AUC and 0.192 in PR AUC over FIB-4 and 0.090 in AUC and 0.130 in PR AUC over APRI. XGBoost also achieved higher recall (0.646 vs 0.564 for FIB-4 and 0.585 for APRI), precision (0.375 vs 0.220 and 0.268), and F1-score (0.474 vs 0.316 and 0.368).

A similar pattern was observed for the 2-year prediction window. XGBoost achieved an AUC of 0.811 and a PR AUC of 0.434, compared with 0.677 and 0.241 for FIB-4 and 0.719 and 0.306 for APRI. This corresponded to improvements of 0.134 in AUC and 0.193 in PR AUC over FIB-4 and 0.092 in AUC and 0.128 in PR AUC over APRI. XGBoost also achieved higher recall (0.655 vs 0.604 for FIB-4 and 0.537 for APRI), precision (0.310 vs 0.189 and 0.267), and F1-score (0.421 vs 0.288 and 0.356). Although predictive performance was somewhat lower at the 2-year horizon than at the 1-year horizon, the relative advantage of XGBoost was preserved.

The ROC and precision–recall curves further demonstrate the consistent advantage of XGBoost over FIB-4 and APRI (Figure 5). In the ROC analyses, XGBoost showed better discrimination across both prediction horizons. The separation was also evident in the precision–recall curves, indicating improved identification of patients who later developed LC in an imbalanced prediction setting. This improvement is particularly important for early risk stratification, where the goal is not only to distinguish cases from controls but also to prioritize patients most likely to benefit from closer monitoring or referral.

Table 4: Model performance for XGBoost, FIB-4, and APRI across 1- and 2-year prediction windows

| Model | AUC | PR_AUC | Precision | Recall | F1 |
|---|---|---|---|---|---|
| One-year XGBoost | 0.834 | 0.502 | 0.375 | 0.646 | 0.474 |
| One-year FIB-4 | 0.7 | 0.31 | 0.22 | 0.564 | 0.316 |
| One-year APRI | 0.744 | 0.372 | 0.268 | 0.585 | 0.368 |
| Two-year XGBoost | 0.811 | 0.434 | 0.31 | 0.655 | 0.421 |
| Two-year FIB-4 | 0.677 | 0.241 | 0.189 | 0.604 | 0.288 |
| Two-year APRI | 0.719 | 0.306 | 0.267 | 0.537 | 0.356 |

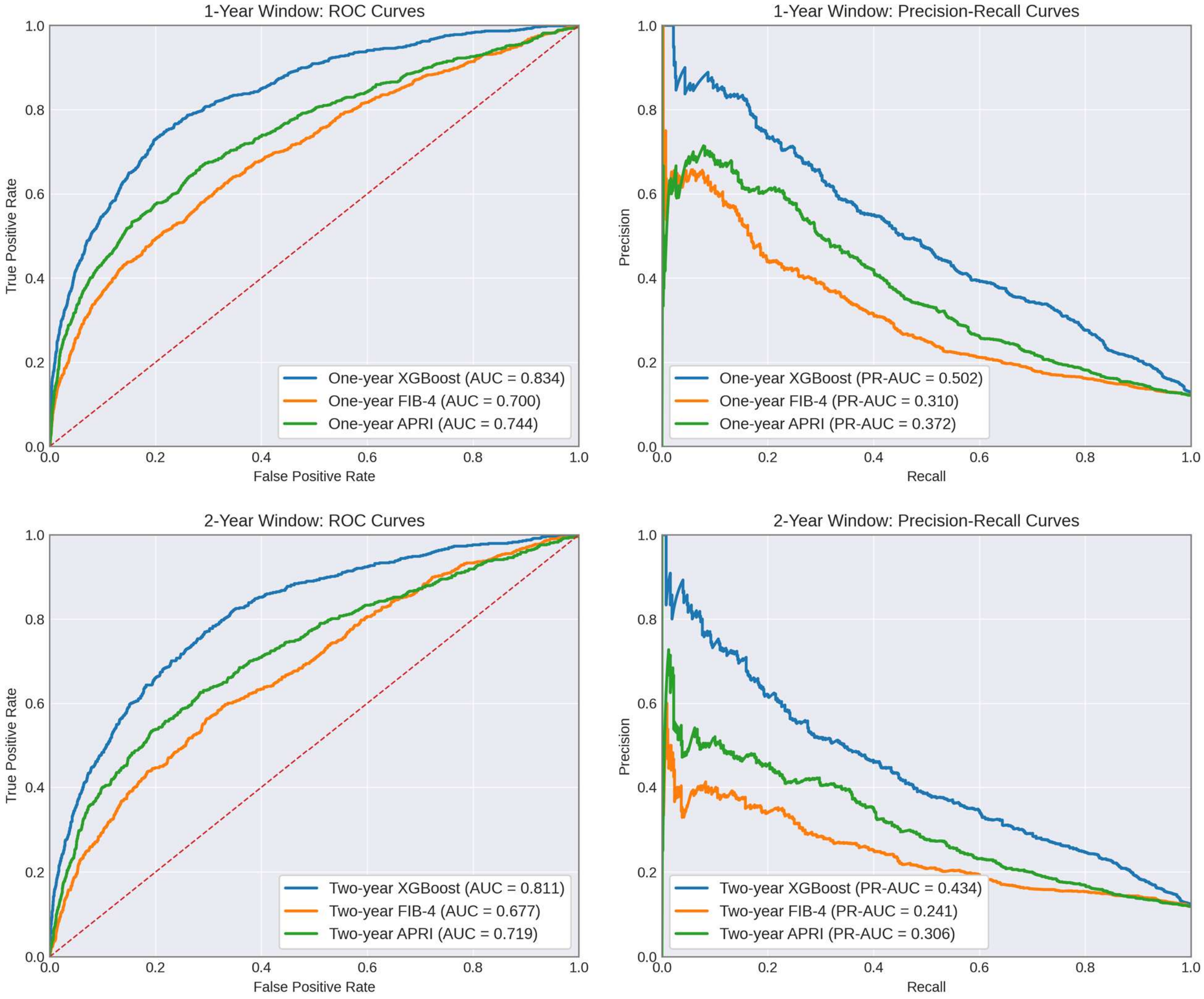

Figure 5. ROC and precision-recall curves for 1- and 2-year predictions from XGBoost models, FIB-4 and APRI scores

The feature selection process retained 36 predictors for the 1-year model and 35 predictors for the 2-year model. Feature importance results are shown in Figure 6 for the 1-year model and Figure A2 in the Appendix for the 2-year model. Across both prediction horizons, the highest-ranked predictors reflected a clinically meaningful combination of liver-related diagnosis indicators, laboratory markers of hepatic injury and function, metabolic risk factors, and demographic characteristics. AST was the highest-ranked predictor in both models, followed by chronic hepatitis C. Other influential predictors shared across the prediction horizons included dyslipidemia/hyperlipidemia, sex, MASLD/MASH, rural-urban status, age, obesity, chronic hepatitis B, alkaline phosphatase, and alcohol use disorder. These findings indicate that the XGBoost models did not rely solely on laboratory values, but instead integrated etiologic, metabolic, biochemical, and demographic information to support early prediction of LC onset.

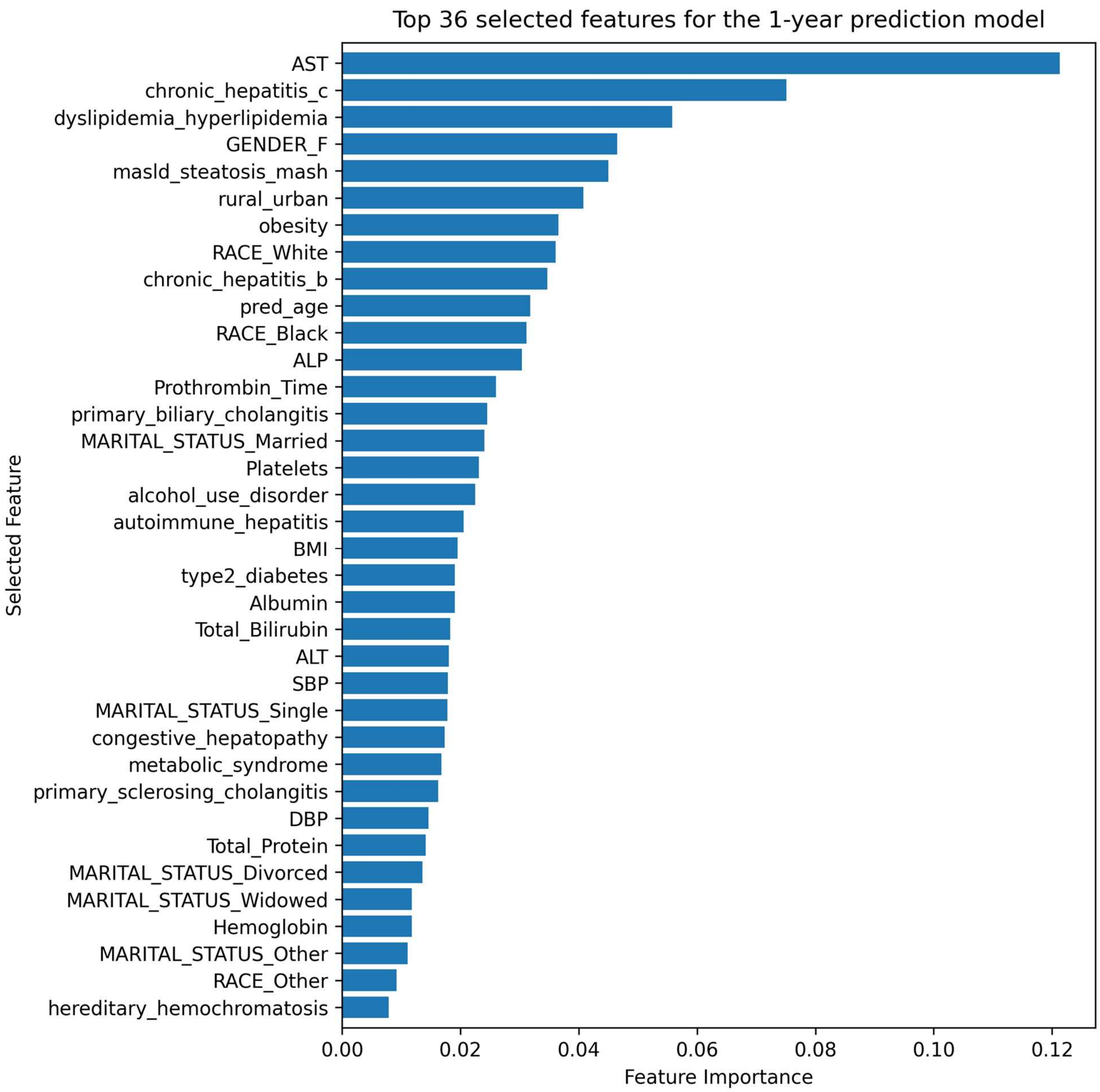

Figure 6: Top 36 features selected for the 1-year prediction model

## Discussion

In this study, we demonstrate that ML–based prediction models substantially outperform the commonly used FIB-4 and APRI scores for identifying patients at risk of developing liver cirrhosis one and two years prior to diagnosis. Across both prediction windows, the XGBoost models achieved consistently higher discriminative performance. Compared with FIB-4, XGBoost improved AUC by 0.134 at both the 1-year and 2-year prediction horizons; compared with APRI, AUC improved by 0.090 and 0.092, respectively. Improvements were also pronounced for PR AUC, with gains of 0.192 and 0.193 over FIB-4 and 0.130 and 0.128 over APRI at the 1- and 2-year horizons, respectively. These findings highlight the limitations of traditional rule-based indices and underscore the potential of ML approaches to improve early risk stratification for cirrhosis in patients with LC-related risk conditions.

The superior performance of the ML models is not unexpected given the fundamental differences in model design. FIB-4 and APRI use predefined formulas based on a limited set of clinical variables and do not learn relationships or interactions directly from the data. While this simplicity enables ease of use, it constrains predictive accuracy and limits adaptability across heterogeneous populations. In contrast, the XGBoost models integrate a wide range of demographic characteristics, clinically grouped diagnosis indicators, laboratory measures, and vital signs, capturing complex relationships and higher-order interactions that more closely reflect the complex, multifactorial pathophysiology of cirrhosis progression. This ability to leverage longitudinal EHR data and model subtle patterns likely explains the improved discrimination observed, particularly in earlier prediction windows where clinical signals may be less pronounced.

The performance advantage of XGBoost was evident across both the 1-year and 2-year prediction windows. Although AUC and PR AUC were modestly lower at the 2-year horizon, XGBoost maintained higher AUC, PR AUC, recall, precision, and F1-score than both FIB-4 and APRI at both prediction horizons. This suggests that the ML approach can preserve useful early risk discrimination even when prediction is performed farther in advance of diagnosis. From a clinical perspective, this finding is important because identifying patients one to two years before LC diagnosis creates a practical window for closer monitoring, lifestyle and metabolic risk-factor modification, and timely hepatology referral before cirrhosis is clinically documented.

The feature importance findings provide additional support for the clinical relevance of the modeling approach. The selected predictors reflected a coherent combination of liver-related etiologies, laboratory markers of hepatic injury and function, metabolic risk factors, and demographic characteristics. AST and chronic hepatitis C were the two highest-ranked predictors across both horizons. Other influential predictors shared across the models included dyslipidemia/hyperlipidemia, sex, MASLD/MASH, rural-urban status, age, obesity, alkaline phosphatase, chronic hepatitis B, and alcohol use disorder. These findings suggest that the models were not driven by laboratory values alone but instead captured the multifactorial nature of LC development by integrating etiologic, metabolic, biochemical, and demographic information.

An important practical advantage of the ML approach is its ability to use incomplete real-world EHR data. FIB-4 and APRI require specific laboratory values and cannot be calculated when the necessary components are missing. In contrast, XGBoost can accommodate missing predictor values during model training and prediction. In routine clinical care, where laboratory testing is often incomplete or irregular, this flexibility increases the feasibility of automated EHR-based risk assessment. With further validation, these models could potentially be implemented as automated processes within the EHR to update patient-level predictions as new clinical information becomes available.

Integration of ML predictions into care workflows also enables more actionable clinical decision support. Rather than providing a binary rule-out or rule-in classification, ML models can generate patient-level predicted probabilities that, following appropriate calibration and clinical validation, could support tiered care pathways. For example, low-risk patients could continue routine monitoring in primary care, intermediate-risk patients could receive targeted laboratory follow-up or lifestyle counseling, and high-risk patients could be flagged automatically for hepatology referral or advanced testing. Such stratification aligns well with population health and value-based care models, where efficient allocation of specialist resources is essential.

Moreover, the use of EHR-based ML may support greater scalability of cirrhosis risk assessment. Because the model uses routinely collected structured EHR information, it could potentially extend risk assessment beyond settings where specialized liver testing is readily available. This is particularly relevant given the observed associations between rural-urban status and cirrhosis risk in our cohort. However, whether such models improve equity or reduce geographic disparities requires dedicated evaluation across different patient populations and healthcare settings.

Several limitations should be acknowledged. First, this study was conducted using data from a single academic health system, which may limit generalizability. External validation across multiple institutions and patient populations will be necessary before widespread implementation. Second, although tree-based models such as XGBoost offer strong performance, their complexity can limit interpretability. Future work should explore explainability techniques to enhance clinician trust and facilitate clinical adoption. Third, the current models used structured EHR variables only; incorporating unstructured clinical notes, imaging findings, medication history, and longitudinal trajectories may further improve prediction. Finally, prospective studies are needed to evaluate whether integrating ML-based risk prediction into clinical workflows leads to improved patient outcomes compared with existing screening strategies.

# Conclusion

This study demonstrates that ML models using routinely collected EHR data can improve early prediction of liver cirrhosis up to two years before diagnosis. Across both the 1-year and 2-year prediction horizons, the optimized XGBoost models outperformed FIB-4 and APRI in terms of AUC, PR AUC, recall, precision, and F1-score when evaluated on the same test subset. By integrating demographics, clinically grouped diagnosis indicators, laboratory results, and vital signs, the ML approach captured a broader and more clinically informative risk profile than traditional serum-based scores. Importantly, the ability of XGBoost to use incomplete real-world EHR data further supports its potential value for automated risk assessment in routine care settings. These findings suggest that EHR-based ML models could support earlier identification of high-risk patients, facilitate closer monitoring and timely hepatology referral, and help shift cirrhosis care from late-stage detection toward earlier, population-level prevention and management.

## References

1. Gan C, Yuan Y, Shen H, et al. Liver diseases: epidemiology, causes, trends and predictions. *Signal Transduct Target Ther*. 2025;10(1):33. doi:10.1038/s41392-024-02072-z

2. Vaz J, Strömberg U, Midlöv P, Eriksson B, Buchebner D, Hagström H. Unrecognized liver cirrhosis is common and associated with worse survival in hepatocellular carcinoma: A nationwide cohort study of 3473 patients. *J Intern Med*. 2023;293(2):184-199. doi:10.1111/joim.13570

3. Flamm SL. Key Insights and Clinical Pearls in the Identification and Management of Cirrhosis and Its Complications. *Am J Med*. 2024;137(10):929-938. doi:10.1016/j.amjmed.2024.05.015

4. Lee S, Saffo S. Evolution of care in cirrhosis: Preventing hepatic decompensation through pharmacotherapy. *World J Gastroenterol*. 2023;29(1):61-74. doi:10.3748/wjg.v29.i1.61

5. Villanueva C, Tripathi D, Bosch J. Preventing the progression of cirrhosis to decompensation and death. *Nat Rev Gastroenterol Hepatol*. 2025;22(4):265-280. doi:10.1038/s41575-024-01031-x

6. Sumida Y, Nakajima A, Itoh Y. Limitations of liver biopsy and non-invasive diagnostic tests for the diagnosis of nonalcoholic fatty liver disease/nonalcoholic steatohepatitis. *World J Gastroenterol*. 2014;20(2):475-485. doi:10.3748/wjg.v20.i2.475

7. Neuberger J, Cain O. The Need for Alternatives to Liver Biopsies: Non-Invasive Analytics and Diagnostics. *Hepatic Med Evid Res*. 2021;13:59-69. doi:10.2147/HMER.S278076

8. Tsoneva DK, Ivanov MN, Vinciguerra M. Liquid Liver Biopsy for Disease Diagnosis and Prognosis. *J Clin Transl Hepatol*. 2023;11(7):1520-1541. doi:10.14218/JCTH.2023.00040

9. Castéra L, Vergniol J, Foucher J, et al. Prospective comparison of transient elastography, Fibrotest, APRI, and liver biopsy for the assessment of fibrosis in chronic hepatitis C. *Gastroenterology*. 2005;128(2):343-350. doi:10.1053/j.gastro.2004.11.018

10. Vallet-Pichard A, Mallet V, Nalpas B, et al. FIB-4: An inexpensive and accurate marker of fibrosis in HCV infection. comparison with liver biopsy and fibrotest. *Hepatology*. 2007;46(1):32-36. doi:10.1002/hep.21669

11. Kumari B, Kumar R, Sharma S, et al. Diagnostic Accuracy of FIB-4 and FIB-5 Scores as Compared to Fibroscan for Assessment of Liver Fibrosis in Patients With Non-Alcoholic Fatty Liver Disease. *Cureus*. Published online August 31, 2021. doi:10.7759/cureus.17622

12. Metwally K, Elsabaawy M, Abdel-Samiee M, Morad W, Ehsan N, Abdelsameea E. FIB-5 versus FIB-4 index for assessment of hepatic fibrosis in chronic hepatitis B affected patients. *Clin Exp Hepatol*. 2020;6(4):335-338. doi:10.5114/ceh.2020.102157

13. Deng H, Qi X, Guo X. Diagnostic Accuracy of APRI, AAR, FIB-4, FI, King, Lok, Forns, and FibroIndex Scores in Predicting the Presence of Esophageal Varices in Liver Cirrhosis: A Systematic Review and Meta-Analysis. *Medicine (Baltimore)*. 2015;94(42):e1795. doi:10.1097/MD.0000000000001795

14. Lee J, Vali Y, Boursier J, et al. Prognostic accuracy of FIB-4, NAFLD fibrosis score and APRI for NAFLD-related events: A systematic review. *Liver Int*. 2021;41(2):261-270. doi:10.1111/liv.14669

15. Tham EKJ, Tan DJH, Danpanichkul P, et al. The Global Burden of Cirrhosis and Other Chronic Liver Diseases in 2021. *Liver Int*. 2025;45(3):e70001. doi:10.1111/liv.70001

16. Ng K, Steinhubl SR, deFilippi C, Dey S, Stewart WF. Early Detection of Heart Failure Using Electronic Health Records. *Circ Cardiovasc Qual Outcomes*. Published online November 2016. Accessed September 25, 2018. https://www.ahajournals.org/doi/abs/10.1161/CIRCOUTCOMES.116.002797

17. Wang R, Miao Z, Liu T, et al. Derivation and Validation of Essential Predictors and Risk Index for Early Detection of Diabetic Retinopathy Using Electronic Health Records. *J Clin Med*. 2021;10(7):7. doi:10.3390/jcm10071473

18. Rural-Urban Continuum Codes | Economic Research Service. Accessed December 18, 2025. https://www.ers.usda.gov/data-products/rural-urban-continuum-codes

19. Chen T, Guestrin C. XGBoost: A Scalable Tree Boosting System. In: *Proceedings of the 22nd ACM SIGKDD International Conference on Knowledge Discovery and Data Mining*. KDD '16. Association for Computing Machinery; 2016:785-794. doi:10.1145/2939672.2939785

20. Quan H, Sundararajan V, Halfon P, et al. Coding Algorithms for Defining Comorbidities in ICD-9-CM and ICD-10 Administrative Data. *Med Care*. 2005;43(11):1130. doi:10.1097/01.mlr.0000182534.19832.83

21. Kansal A, Gao M, Balu S, et al. Impact of diagnosis code grouping method on clinical prediction model performance: A multi-site retrospective observational study. *Int J Med Inf*. 2021;151:104466. doi:10.1016/j.ijmedinf.2021.104466

22. Malecki SL, Loffler A, Tamming D, et al. Development and external validation of tools for categorizing diagnosis codes in international hospital data. *Int J Med Inf*. 2024;189:105508. doi:10.1016/j.ijmedinf.2024.105508

23. Kumar D, Sood SK, Rawat KS. Early health prediction framework using XGBoost ensemble algorithm in intelligent environment. *Artif Intell Rev*. 2023;56(1):1591-1615. doi:10.1007/s10462-023-10565-6

# Appendix

Table A1. Cohort characteristics for the 2-year prediction

| | Overall | Controls | Cases | p-value |
|---|---|---|---|---|
| **Patient Counts** | 43,743 | 38,325 | 5,418 | |
| **Gender** | | | | <0.001 |
| Female | 25,839 (59.1%) | 23,335 (60.9%) | 2,504 (46.2%) | |
| Male | 17,904 (40.9%) | 14,990 (39.1%) | 2,914 (53.8%) | |
| **Race** | | | | <0.001 |
| White | 24,680 (56.4%) | 21,114 (55.1%) | 3,566 (65.8%) | |
| Black | 16,682 (38.1%) | 15,072 (39.3%) | 1,610 (29.7%) | |
| Other | 2,381 (5.4%) | 2,139 (5.6%) | 242 (4.5%) | |
| **Marital Status** | | | | <0.001 |
| Married | 20,757 (47.5%) | 18,409 (48.0%) | 2,348 (43.3%) | |
| Single | 12,501 (28.6%) | 10,887 (28.4%) | 1,614 (29.8%) | |
| Divorced | 4,537 (10.4%) | 3,844 (10.0%) | 693 (12.8%) | |
| Widowed | 4,229 (9.7%) | 3,735 (9.7%) | 494 (9.1%) | |
| Other | 1,719 (3.9%) | 1,450 (3.8%) | 269 (5.0%) | |
| **Rural-Urban Status** | | | | <0.001 |
| Metro - Counties in metro areas of 1 million population or more | 33,022 (75.5%) | 29,562 (77.1%) | 3,460 (63.9%) | |
| Metro - Counties in metro areas of 250,000 to 1 million population | 3,303 (7.6%) | 2,699 (7.0%) | 604 (11.1%) | |
| Metro - Counties in metro areas of fewer than 250,000 population | 2,841 (6.5%) | 2,328 (6.1%) | 513 (9.5%) | |
| Nonmetro - Urban population of 20,000 or more, adjacent to a metro area | 2,117 (4.8%) | 1,746 (4.6%) | 371 (6.8%) | |
| Nonmetro - Urban population of 20,000 or more, not adjacent to a metro area | 189 (0.4%) | 155 (0.4%) | 34 (0.6%) | |
| Nonmetro - Urban population of 5,000 to 20,000, adjacent to a metro area | 955 (2.2%) | 774 (2.0%) | 181 (3.3%) | |
| Nonmetro - Urban population of 5,000 to 20,000, not adjacent to a metro area | 108 (0.2%) | 81 (0.2%) | 27 (0.5%) | |
| Nonmetro - Urban population of fewer than 5,000, adjacent to a metro area | 773 (1.8%) | 623 (1.6%) | 150 (2.8%) | |
| Nonmetro - Urban population of fewer than 5,000, not adjacent to a metro area | 435 (1.0%) | 357 (0.9%) | 78 (1.4%) | |
| **Age, mean (SD)** | 54.6 (15.9) | 54.3 (16.2) | 57.0 (13.3) | <0.001 |
| **Vital and Labs, mean (SD)** | | | | |
| BMI | 31.6 (7.8) | 31.8 (7.8) | 30.7 (7.8) | <0.001 |
| DBP | 79.2 (8.4) | 79.1 (8.3) | 79.9 (9.3) | <0.001 |
| SBP | 132.1 (15.2) | 131.7 (14.9) | 134.9 (16.9) | <0.001 |
| ALT | 25.6 (32.1) | 23.6 (26.6) | 40.5 (56.7) | <0.001 |
| Albumin | 4.0 (0.4) | 4.1 (0.4) | 3.9 (0.5) | <0.001 |
| AST | 26.2 (30.2) | 23.9 (26.6) | 43.4 (45.9) | <0.001 |
| Bilirubin | 0.6 (0.5) | 0.6 (0.4) | 0.9 (1.0) | <0.001 |
| Platelets | 236.8 (76.6) | 240.5 (74.4) | 209.3 (86.1) | <0.001 |
| Total Protein | 7.0 (0.6) | 7.0 (0.6) | 7.1 (0.7) | <0.001 |
| ALP | 80.0 (44.6) | 76.8 (32.3) | 104.1 (91.1) | <0.001 |
| Hemoglobin | 13.1 (1.7) | 13.1 (1.7) | 13.0 (1.9) | 0.522 |
| **LC-related risk indicators, Present, n(%)** | | | | |
| Alcohol-related liver disease | 105 (0.2%) | 35 (0.1%) | 70 (1.3%) | <0.001 |
| Alcohol use disorder | 1,901 (4.3%) | 1,380 (3.6%) | 521 (9.6%) | <0.001 |
| Alpha1 antitrypsin deficiency | 58 (0.1%) | 42 (0.1%) | 16 (0.3%) | <0.001 |
| Autoimmune hepatitis | 105 (0.2%) | 38 (0.1%) | 67 (1.2%) | <0.001 |
| Chronic hepatitis b | 199 (0.5%) | 98 (0.3%) | 101 (1.9%) | <0.001 |
| Chronic hepatitis c | 867 (2.0%) | 464 (1.2%) | 403 (7.4%) | <0.001 |
| Congestive hepatopathy | 92 (0.2%) | 59 (0.2%) | 33 (0.6%) | <0.001 |
| Drug-induced liver injury | 104 (0.2%) | 71 (0.2%) | 33 (0.6%) | <0.001 |
| dyslipidemia hyperlipidemia | 15,828 (36.2%) | 14,447 (37.7%) | 1,381 (25.5%) | <0.001 |
| Hereditary hemochromatosis | 20 (0.0%) | 13 (0.0%) | 7 (0.1%) | 0.006 |
| Masld-steatosis-mash | 2,236 (5.1%) | 1,658 (4.3%) | 578 (10.7%) | <0.001 |
| Metabolic syndrome | 765 (1.7%) | 699 (1.8%) | 66 (1.2%) | 0.002 |
| Obesity | 14,600 (33.4%) | 13,317 (34.7%) | 1,283 (23.7%) | <0.001 |
| Primary biliary cholangitis | 77 (0.2%) | 13 (0.0%) | 64 (1.2%) | <0.001 |
| Primary sclerosing cholangitis | 86 (0.2%) | 21 (0.1%) | 65 (1.2%) | <0.001 |
| Toxic exposure | 160 (0.4%) | 150 (0.4%) | 10 (0.2%) | 0.025 |
| Type2_diabetes | 10,625 (24.3%) | 9,258 (24.2%) | 1,367 (25.2%) | 0.087 |
| Wilson disease | 6 (0.0%) | 4 (0.0%) | 2 (0.0%) | 0.348 |

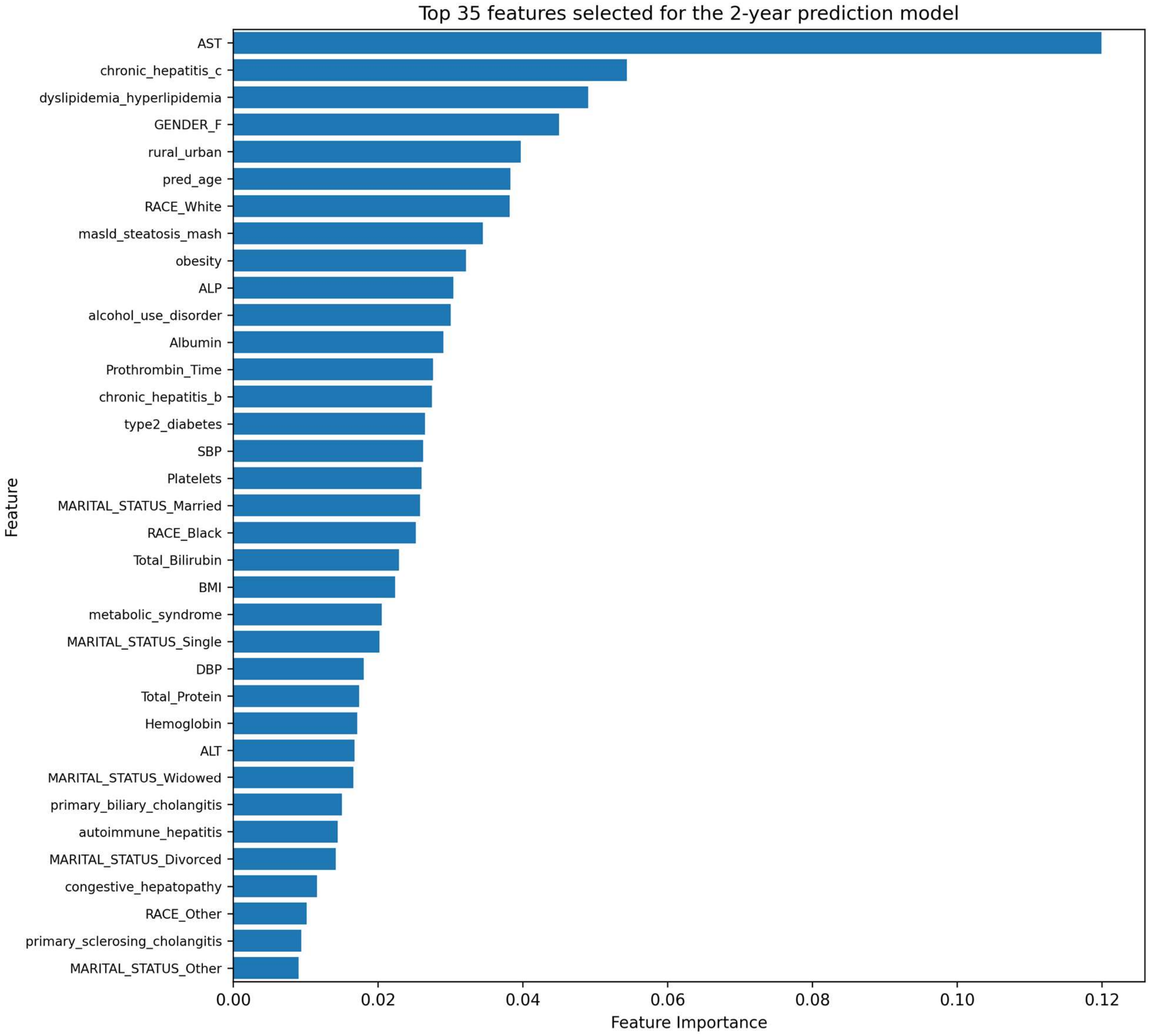

Figure A1: Top 35 features selected for the 2-year prediction model